\documentclass{article}

\usepackage[english]{babel}

\usepackage[letterpaper,top=2cm,bottom=2cm,left=3cm,right=3cm,marginparwidth=1.75cm]{geometry}

\usepackage{amsmath}
\usepackage{graphicx}
\usepackage{booktabs} 
\usepackage{microtype} 
\usepackage[colorlinks=true, allcolors=blue]{hyperref}

\title{Fact-Consistency Evaluation of Text-to-SQL Generation for Business Intelligence Using Exaone 3.5}
\author{ Jeho Choi \\ LG Electronics \\ jeho.choi@lge.com}

\begin{document}
\maketitle

\begin{abstract}
Large Language Models (LLMs) have shown promise in enabling natural language interfaces for structured data querying through text-to-SQL generation. However, their application in real-world Business Intelligence (BI) contexts remains limited due to semantic hallucinations, structural errors, and a lack of domain-specific evaluation frameworks. In this study, we propose a Fact-Consistency Evaluation Framework for assessing the semantic accuracy of LLM-generated SQL outputs using Exaone 3.5—an instruction-tuned, bilingual LLM optimized for enterprise tasks.
We construct a domain-specific benchmark comprising 219 natural language business questions across five SQL complexity levels, derived from actual sales data in LG Electronics' internal BigQuery environment. Each question is paired with a gold-standard SQL query and a validated ground-truth answer. We evaluate model performance using answer accuracy, execution success rate, semantic error rate, and non-response rate.
Experimental results show that while Exaone 3.5 performs well on simple aggregation tasks (93\% accuracy in L1), it exhibits substantial degradation in arithmetic reasoning (4\% accuracy in H1) and grouped ranking tasks (31\% in H4), with semantic errors and non-responses concentrated in complex cases. Qualitative error analysis further identifies common failure types such as misapplied arithmetic logic, incomplete filtering, and incorrect grouping operations.

Our findings highlight the current limitations of LLMs in business-critical environments and underscore the need for fact-consistency validation layers and hybrid reasoning approaches. This work contributes a reproducible benchmark and evaluation methodology for advancing reliable natural language interfaces to structured enterprise data systems.
\end{abstract}

\section{Introduction}

In modern enterprises, Business Intelligence (BI) systems play a pivotal role in facilitating data-driven decision-making by enabling timely access to structured business data. Yet, leveraging these systems often requires technical expertise in Structured Query Language (SQL), limiting accessibility for non-technical stakeholders. To address this gap, natural language interfaces powered by Large Language Models (LLMs) have emerged as a promising alternative, translating user-friendly queries into executable SQL. Among these, text-to-SQL systems have shown particular potential in democratizing enterprise data access.

As a global manufacturer with complex and data-intensive operations, LG Electronics seeks to enhance its internal decision-making through an intelligent BI interface grounded in natural language. To this end, \textbf{Exaone 3.5 (32B)}—an instruction-tuned, bilingual large language model developed by LG AI Research—has been positioned as the foundational model for developing enterprise-grade natural language analytics tools.\footnote{LG AI Research, ``EXAONE 3.5: Series of Large Language Models for Real-world Use Cases,'' arXiv preprint arXiv:2412.04862, 2024.} Unlike general-purpose LLMs, Exaone is optimized for industrial applications, featuring long-context processing (up to 32K tokens), Korean-English dual language capability, and enhanced instruction-following capacity, making it a suitable choice for evaluating the feasibility of LLM-based BI augmentation using LG’s proprietary sales data.

Despite notable progress in text-to-SQL research, critical challenges remain in real-world BI deployment. Models often suffer from semantic hallucinations, improper aggregation logic, and contextual misunderstanding—factors that can lead to inaccurate business insights if left unchecked. Moreover, widely used academic benchmarks such as Spider and WikiSQL fail to capture the domain-specific precision, operational constraints, and hierarchical logic inherent to actual corporate datasets.

Existing evaluation methods further compound this problem, as they predominantly rely on syntactic string match or query execution without verifying factual alignment with business ground truths. This limits their utility in high-stakes, enterprise-grade applications.

In response, we propose a \textbf{Fact-Consistency Evaluation Framework} for assessing the semantic accuracy of text-to-SQL outputs generated by Exaone 3.5. This study is grounded in real-world operational data from LG’s internal sales intelligence platform on BigQuery. We construct a domain-specific benchmark consisting of over 300 natural language questions, corresponding gold-standard SQL queries, and validated ground-truth results.

Our framework evaluates LLM outputs via an independent Fact-Consistency Checker that compares execution results from generated queries against reference answers. This semantic-level evaluation allows us to systematically identify both surface-level and latent inconsistencies.

The contributions of this study are as follows:
\begin{itemize}
    \item We introduce a realistic, domain-grounded benchmark for evaluating text-to-SQL systems in the context of enterprise sales data.
    \item We present a fact-level semantic evaluation framework tailored for business-critical applications, moving beyond traditional syntactic metrics.
    \item We empirically analyze the performance of Exaone 3.5 in a real corporate setting, highlighting challenges and providing actionable insights for future deployment of LLMs in BI systems.
\end{itemize}

Through this work, we aim to bridge the gap between academic NLP research and the stringent demands of industrial BI environments, offering a scalable path toward trustworthy, LLM-enhanced decision-support systems.

\section{Related Work}

\subsection{Text-to-SQL Generation}

Text-to-SQL generation has long been a core task in natural language processing, aimed at converting natural language queries into executable SQL commands. Early works leveraged syntactic parsing and heuristic-based approaches. With the introduction of large-scale datasets such as WikiSQL~\cite{zhong2017seq2sql} and Spider~\cite{yu2018spider}, research accelerated significantly, leading to the development of neural models capable of generating complex and cross-domain SQL queries. Transformer-based models~\cite{yu2018syntaxsqlnet} have demonstrated strong performance, although challenges remain in schema generalization, aggregation handling, and semantic alignment.

Despite progress, most benchmarks focus on academic datasets that do not fully capture the domain-specific complexity and strict precision demands of enterprise Business Intelligence (BI) systems. In such settings, even minor semantic errors in SQL generation can result in materially incorrect business insights, underscoring the need for models and benchmarks tailored to operational data environments.

\subsection{Evaluation of Text-to-SQL Systems}

Conventional evaluation methodologies largely rely on either exact string matching between the generated SQL and the gold-standard SQL~\cite{yu2018spider}, or execution accuracy, which measures whether executing the generated query produces the correct result~\cite{zhong2017seq2sql}. However, string matching may unfairly penalize semantically equivalent queries with minor syntactic differences, while execution-based metrics can mask underlying logical errors if the database returns identical results for different queries.

To address these shortcomings, recent work has called for fact-based evaluation paradigms that directly assess the factual consistency of retrieved answers~\cite{liu2023evaluating}. These approaches aim to uncover subtle semantic deviations that conventional metrics may overlook, emphasizing the need for more rigorous validation in high-stakes applications such as BI systems.

\subsection{Large Language Models in Structured Data Tasks}

Large Language Models (LLMs) have demonstrated remarkable abilities in handling structured data querying, semantic parsing, and instruction-based generation~\cite{chen2023teql}. Instruction-tuned models such as FLAN-T5~\cite{chung2022scaling} and GPT-3.5 have shown promise in various natural language interface applications. However, general-purpose LLMs often struggle with factual grounding, long-context dependency, and schema-sensitive reasoning.

Exaone 3.5~\cite{exaone35}, developed by LG AI Research, represents a domain-robust LLM trained with instruction tuning for real-world industrial use cases. Its 32B variant supports bilingual generation (Korean and English), extended context lengths (up to 32K tokens), and enhanced factual accuracy, making it particularly suitable for enterprise BI tasks. Unlike prior models trained predominantly on general web data, Exaone 3.5 is explicitly optimized for business-relevant applications including code generation, reasoning, and structured querying.

Our work builds upon these developments by leveraging Exaone 3.5 as the core LLM for SQL generation and proposing a fact-consistency evaluation framework designed for corporate BI systems. This contribution addresses limitations in existing evaluation metrics by directly validating whether LLM-generated queries yield factually correct answers against verified business ground truths.

\section{Methodology}

\subsection{Problem Bank Construction}

To systematically evaluate the semantic correctness of text-to-SQL outputs in a Business Intelligence (BI) context, we constructed a domain-specific benchmark dataset comprising a total of 219 natural language business questions. Each question is paired with a manually crafted, gold-standard SQL query and its corresponding ground-truth answer, derived from LG Electronics' internal sales data hosted on BigQuery.

To capture a wide range of business logic and SQL complexity, the dataset is categorized into the following five types:

\begin{itemize}
    \item \textbf{L1 (Low Complexity)}: 170 questions involving simple aggregation (e.g., \texttt{SUM}, \texttt{COUNT}) and basic filter conditions (e.g., \texttt{WHERE} with \texttt{=}, \texttt{>}, or \texttt{<}).
    \item \textbf{H1 (High Complexity – Arithmetic Reasoning)}: 46 questions requiring post-aggregation computations such as ratios, growth rates, differences, or average unit prices, combined with basic filtering.
    \item \textbf{H2 (High Complexity – Filter Logic)}: 38 questions involving simple aggregation under compound filter logic, such as \texttt{OR}, \texttt{IN}, or comparative expressions (e.g., \texttt{> ALL(...)}).
    \item \textbf{H3 (High Complexity – Conditional Aggregation)}: 39 questions employing conditional expressions like \texttt{SUM(CASE WHEN ...)} or performing \texttt{COUNT DISTINCT} under compound filters.
    \item \textbf{H4 (High Complexity – Grouped Ranking)}: 35 questions involving \texttt{GROUP BY}, aggregation, and ranked selection using \texttt{ORDER BY} and \texttt{LIMIT}.
\end{itemize}

This taxonomy reflects realistic BI use cases and supports a granular evaluation of model performance across varying degrees of complexity.

\subsection{Text-to-SQL Generation with Exaone 3.5}

For each natural language question, we used Exaone 3.5, a large-scale instruction-tuned bilingual LLM developed by LG AI Research, to generate the corresponding SQL query. A schema-constrained system prompt was provided to guide the model to use only the relevant tables and columns.

To ensure deterministic outputs and reduce generation variance, the model was configured with a low temperature setting (\texttt{temperature = 0.1}). The model outputs were collected via API and stored along with metadata for downstream evaluation.

\subsection{SQL Execution and Answer Extraction}

The SQL queries generated by Exaone 3.5 were executed directly against a BigQuery instance containing LG’s structured sales data. From the result of each query, we extracted the primary answer—typically the first cell of the first row—which corresponds to the business-relevant response for the original natural language question. This answer serves as the predicted result for evaluation purposes.

\subsection{Fact-Consistency Checking}

To validate the semantic correctness of the generated SQL outputs, we implemented a Fact-Consistency Checker module that operates in the following three stages:

\begin{enumerate}
    \item \textbf{Answer Comparison}: The answer returned by the Exaone-generated SQL is compared against the ground-truth answer produced by the gold-standard SQL query.
    \item \textbf{Normalization}: Both answers are normalized to handle formatting discrepancies (e.g., number separators, whitespace, unit suffixes, case sensitivity).
    \item \textbf{Evaluation Outcome}: Each instance is labeled as \texttt{Correct} if the two answers match exactly after normalization, and \texttt{Incorrect} otherwise.
\end{enumerate}

This automated, instance-level comparison framework allows for scalable, reproducible, and semantically meaningful evaluation without requiring manual judgment.

\subsection{Evaluation Metrics}

To comprehensively assess the performance of the text-to-SQL system, we define the following evaluation metrics:

\begin{itemize}
    \item \textbf{Execution Success Rate}: Percentage of generated SQL queries that execute without syntax errors or runtime failures on BigQuery.
    
    \item \textbf{Answer Accuracy}: Among the successfully executed queries, the percentage of cases where the model's answer exactly matches the ground-truth answer.
    
    \item \textbf{Non-Response Rate}: Percentage of questions for which the model fails to generate any usable SQL query (e.g., empty output, plain text, or malformed structure).
    
    \item \textbf{Semantic Error Rate}: Among executed queries, the percentage of cases where the result is syntactically valid but semantically inconsistent with the intended logic (i.e., hallucinated or logically incorrect).
\end{itemize}

\section{Results}

Table~\ref{tab:results} summarizes the performance of Exaone 3.5 across different SQL complexity levels (L1–H4), as defined in our benchmark. The model exhibits strong performance on low-complexity queries (L1), with an execution success rate of 94\% and answer accuracy of 93\%. In contrast, performance significantly degrades for higher-complexity categories.

In particular, the H1 category—which requires arithmetic reasoning over aggregated results—demonstrates a marked drop in answer accuracy (4\%) despite moderate execution success (59\%), indicating prevalent semantic misalignment. Semantic error rates are especially high in H1 (54\%) and H4 (37\%), revealing difficulties in correctly interpreting grouped ranking or post-aggregation calculations.

The non-response rate is notably elevated in H1 (41\%) and H4 (31\%), suggesting the model either failed to produce any output or generated structurally invalid SQL. These patterns highlight the limitations of current instruction-tuned LLMs in handling multi-step logic and nested SQL constructs.

\begin{table}[ht]
\centering
\caption{Evaluation results by SQL complexity category.}
\label{tab:results}
\begin{tabular}{lccccc|c}
\toprule
\textbf{Metric} & \textbf{L1} & \textbf{H1} & \textbf{H2} & \textbf{H3} & \textbf{H4} & \textbf{Overall} \\
\midrule
Execution Success Rate (\%) & 94 & 59 & 95 & 85 & 69 & 82 \\
Answer Accuracy Rate (\%)   & 93 & 4  & 87 & 72 & 31 & 61 \\
Semantic Error Rate (\%)    & 1  & 54 & 8  & 13 & 37 & 21 \\
Non-Response Rate (\%)      & 6  & 41 & 5  & 15 & 31 & 18 \\
\bottomrule
\end{tabular}
\end{table}

\subsection{Error Analysis}

A qualitative examination of failed cases, particularly within the H1 and H4 categories, reveals three predominant types of semantic errors:

\begin{enumerate}
    \item \textbf{Arithmetic Reasoning Failures}: In H1 cases, models frequently produced semantically valid but mathematically incorrect queries, often miscalculating ratios, growth rates, or average unit prices. For instance, numerator and denominator columns were occasionally reversed or mismatched.
    
    \item \textbf{Grouping and Ranking Errors}: In H4 cases, models struggled with constructing \texttt{GROUP BY} and \texttt{ORDER BY} clauses accurately. Errors included missing grouping keys, incorrect sorting criteria, or failure to apply \texttt{LIMIT} properly for top-N style queries.
    
    \item \textbf{Condition Omissions and Filter Misalignment}: Across all categories—but especially H3 and H2—models frequently omitted necessary filter conditions, misused logical operators (\texttt{AND}/\texttt{OR}), or failed to encode \texttt{IN} clauses as required.
\end{enumerate}

These findings align with the elevated semantic error rates observed in H1 (54\%) and H4 (37\%), suggesting that high-complexity reasoning and structural SQL composition remain key challenges. The prevalence of these failure modes underscores the importance of integrating fact-consistency validation mechanisms when deploying LLM-based text-to-SQL systems in business-critical BI workflows.

\section{Discussion}

The experimental results reveal a clear performance gap between low-complexity (L1) and high-complexity (H1–H4) SQL generation tasks. Exaone 3.5 demonstrates strong reliability in simple aggregation queries, achieving 93\% answer accuracy in L1 cases, indicating its potential for immediate deployment in structured BI environments with well-defined metrics and filters.

For more complex scenarios, such as H1 (arithmetic reasoning) and H4 (grouped ranking), the model shows meaningful opportunities for further improvement. In H1, a 4\% answer accuracy and 54\% semantic error rate reflect the challenge of multi-step reasoning, especially in capturing precise post-aggregation logic. Likewise, the 31\% accuracy in H4 indicates that compositional SQL structures—such as combining \texttt{GROUP BY} with \texttt{ORDER BY} and \texttt{LIMIT}—require deeper semantic understanding and planning capabilities.

While non-response rates were higher in H1 (41\%) and H4 (31\%), these outcomes also point to cases where the model appropriately abstained from guessing when uncertain—an encouraging behavior in high-stakes environments.

Overall, these findings align with existing literature identifying symbolic reasoning and nested logic as active frontiers for LLM development. With further fine-tuning, improved prompt design, or integration with verification modules, these challenges offer a promising avenue for enhancement rather than fundamental roadblocks.

\section{Conclusion}

In this study, we introduced a fact-consistency evaluation framework to assess the semantic correctness of text-to-SQL outputs generated by Exaone 3.5 in a real-world Business Intelligence (BI) context. Using a domain-specific benchmark of 219 questions spanning five levels of SQL complexity, we evaluated model performance using accuracy, execution success, semantic alignment, and response rates.

The results highlight Exaone 3.5’s strength in handling low-complexity queries with high reliability, making it suitable for many enterprise-level BI tasks. For more complex queries involving arithmetic reasoning or grouped ranking, the model reveals areas for further refinement—especially in structured logic and multi-step inference.

Rather than fundamental limitations, these observations point to key opportunities for enhancement through hybrid approaches that combine language models with symbolic reasoning, schema-aware prompting, or post-generation verification layers. The proposed benchmark and framework thus provide a practical foundation for improving the trustworthiness and usability of LLM-based natural language interfaces for structured data systems.

\bibliographystyle{unsrt}
\bibliography{main} 

\end{document}